\documentclass[10pt,twocolumn,letterpaper]{article}

\usepackage[pagenumbers]{cvpr} % To force page numbers, e.g. for an arXiv version

\usepackage[dvipsnames]{xcolor}

\definecolor{cvprblue}{rgb}{0.21,0.49,0.74}
\usepackage[pagebackref,breaklinks,colorlinks,citecolor=cvprblue]{hyperref}

\title{DreamTuner: Single Image is Enough for Subject-Driven Generation}

\author{
Miao Hua$^*$\quad Jiawei Liu$^*$\quad Fei Ding$^*$ \quad Wei Liu \quad Jie Wu \quad Qian He\\
ByteDance Inc.\\
BeiJing, China\\
\{huamiao,liujiawei.cc22,dingfei.212,liuwei.jikun,wujie.10,heqian\}@bytedance.com\\
\url{https://dreamtuner-diffusion.github.io/}\\
\begin{small}
* Equal Contribution
\end{small}
}

\begin{document}
\twocolumn[{%
\renewcommand\twocolumn[1][]{#1}%
\maketitle
\vspace{-10mm}
\begin{center}
    \centering
    \includegraphics[width=\linewidth]{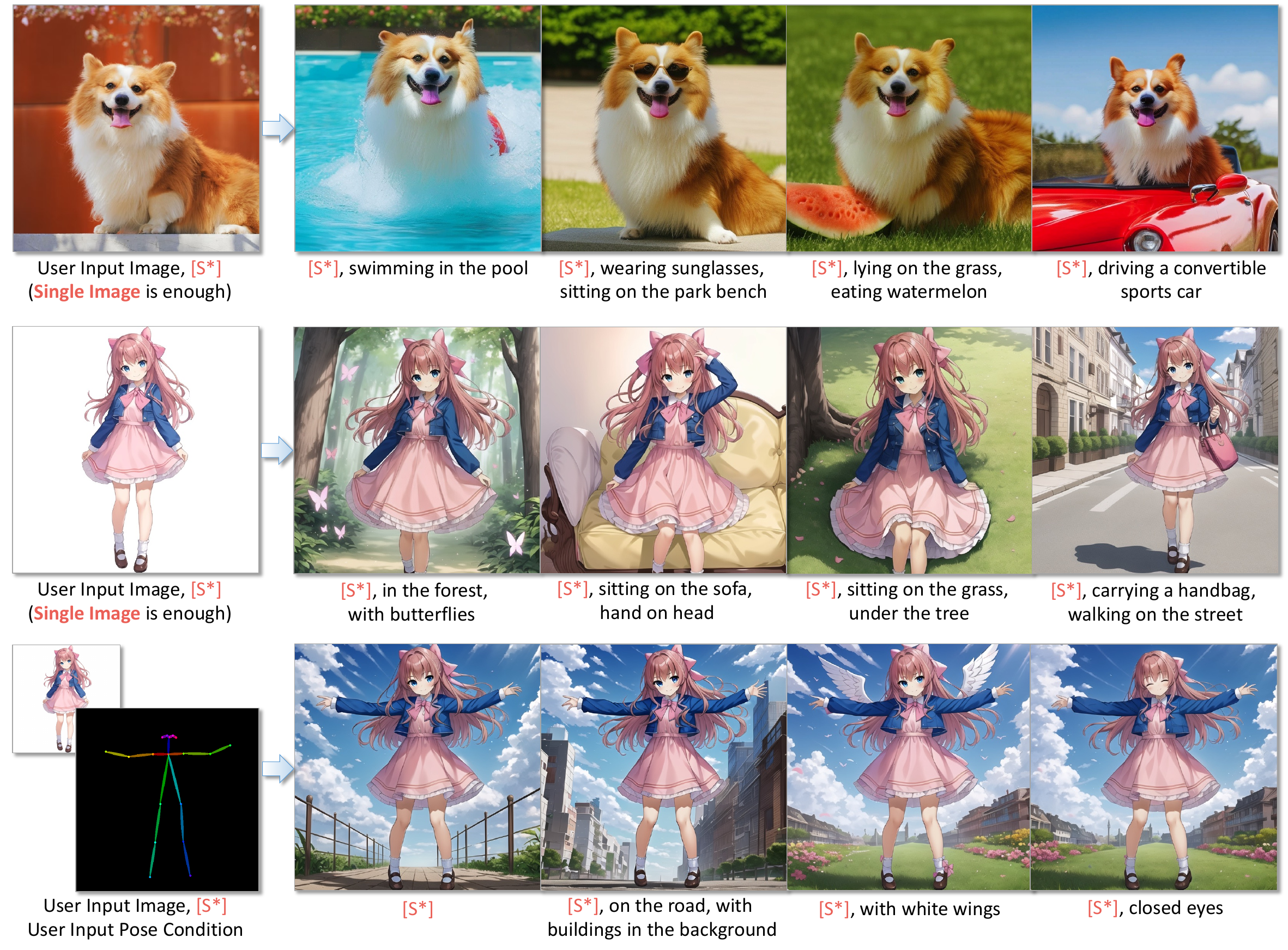}
    \captionof{figure}{Subject-driven image generation results of our method.
  Our proposed DreamTuner could generate high-fidelity images of user input subject, guided by complex texts (the first two rows) or other conditions as pose (the last row), while maintaining the identity appearance of the specific subject. 
  \emph{We found that a single image is enough for surprising subject-driven image generation.}}
    \label{fig:teaser}
\end{center}
}]

\begin{abstract}
Diffusion-based models have demonstrated impressive capabilities for text-to-image generation and are expected for personalized applications of subject-driven generation, which require the generation of customized concepts with one or a few reference images. 
However, existing methods based on fine-tuning fail to balance the trade-off between subject learning and the maintenance of the generation capabilities of pretrained models. 
Moreover, other methods that utilize additional image encoders tend to lose important details of the subject due to encoding compression. 
To address these challenges, we propose DreamTurner\footnote{This work was done in May 2023.}, a novel method that injects reference information from coarse to fine to achieve subject-driven image generation more effectively. 
DreamTurner introduces a subject-encoder for coarse subject identity preservation, where the compressed general subject features are introduced through an attention layer before visual-text cross-attention. We then modify the self-attention layers within pretrained text-to-image models to self-subject-attention layers to refine the details of the target subject. 
The generated image queries detailed features from both the reference image and itself in self-subject-attention. 
It is worth emphasizing that self-subject-attention is an effective, elegant, and training-free method for maintaining the detailed features of customized subjects and can serve as a plug-and-play solution during inference. 
Finally, with additional subject-driven fine-tuning, DreamTurner achieves remarkable performance in subject-driven image generation, which can be controlled by a text or other conditions such as pose. 
% For further details, please visit the project page at \url{https://dreamtuner-diffusion.github.io/}.
\end{abstract}

\section{Introduction}
% 先介绍扩散模型在文生图和编辑都取得出色的效果
Latest developments in diffusion models have shown astonishing achievements in both text-to-image (T2I) generation \cite{glide, dalle2, imagen, ldm} and editing \cite{controlnet, prompt_to_prompt, instruct_pix2pix}.
% 引出后续研究期望实现定制化的subject driven generation，并介绍是什么。
Further researches are expected for personalized applications of subject-driven image generation \cite{textualinversion, dreambooth, imagic, elite,tiue}. 
The goal of subject-driven image generation is to synthesis various high fidelity images of customized concepts, as shown in Fig. \ref{fig:teaser}. 
This is a practical task with a wide range of multimedia applications. For example, merchants can generate advertising images for specific products, and designers can create storybooks by simply drawing an initial character image. When generating images with only a single reference image, it becomes more practical as it reduces user operations and allows the initial reference image to be a generated one.

However, achieving single-reference image subject-driven generation remains a challenging task due to the difficulty in balancing detailed identity preservation and model controllability. Current subject-driven image generation methods are primarily based on fine-tuning or additional image encoders. Fine-tuning-based methods \cite{textualinversion, dreambooth, imagic} train pretrained T2I models on specific reference images to learn the identities of the target subjects. Though more training steps may improve identity preservation, they may also undermine the generation capacity of the pretrained models. Image-encoder-based methods \cite{elite, tiue, tamingencoder} train additional image encoders to inject features of the reference image into the pretrained T2I models. However, these methods require a trade-off of encoding compression. Strong compression may result in the loss of details of the target subject, while weak compression can easily result in the generated image collapsing to a simple reconstruction of the reference image.

\begin{figure}
  \centering
  \includegraphics[width=\linewidth]{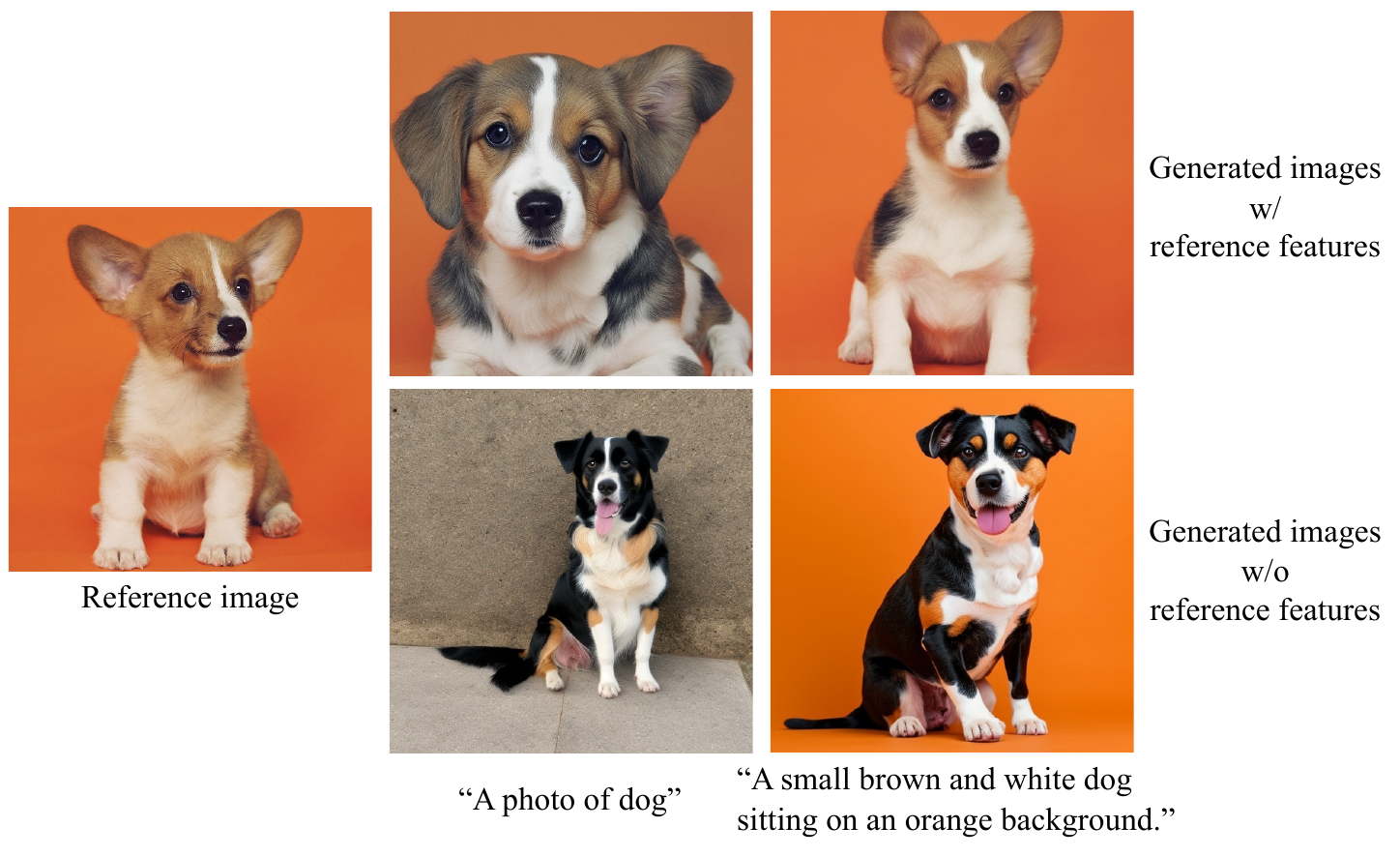}
  \caption{Exploration experiment of self-attention. It makes the generated images more similar to the reference one to use the reference image features for self-attention. Detailed text can better serve its purpose.}
  \label{fig:motivation}
\end{figure}

In fact, as shown in some video generation methods \cite{tune_a_video,t2v_zero}, self-attention layers in T2I models are detailed spatial contextual association modules that could be extended to multiple images for temporal consistency. 
We performed a simple experiment to demonstrate that self-attention layers could also preserve the detailed identity features of the subject, as shown in Fig. \ref{fig:motivation}. 
At each diffusion time step, the reference image is noised through the diffusion process and the reference features could be extracted from the noised image. Then the reference features and those of the generated image are concatenated and used as the key and value of the self-attention layers. 
This approach resembles image outpainting, but it does not modify the convolutional layers and cross-attention layers. Consequently, we found that the generated images with reference features were significantly more similar to the reference image in terms of detailed subject identity preservation, indicating that self-attention plays a role in this aspect. However, as shown in the comparison of generated images guided by detailed and concise text, it is still necessary to introduce coarse identity preservation methods so that the model could generate a rough appearance to better utilize the self-attention layers.

\begin{figure*}
  \centering
  \includegraphics[width=0.9\linewidth]{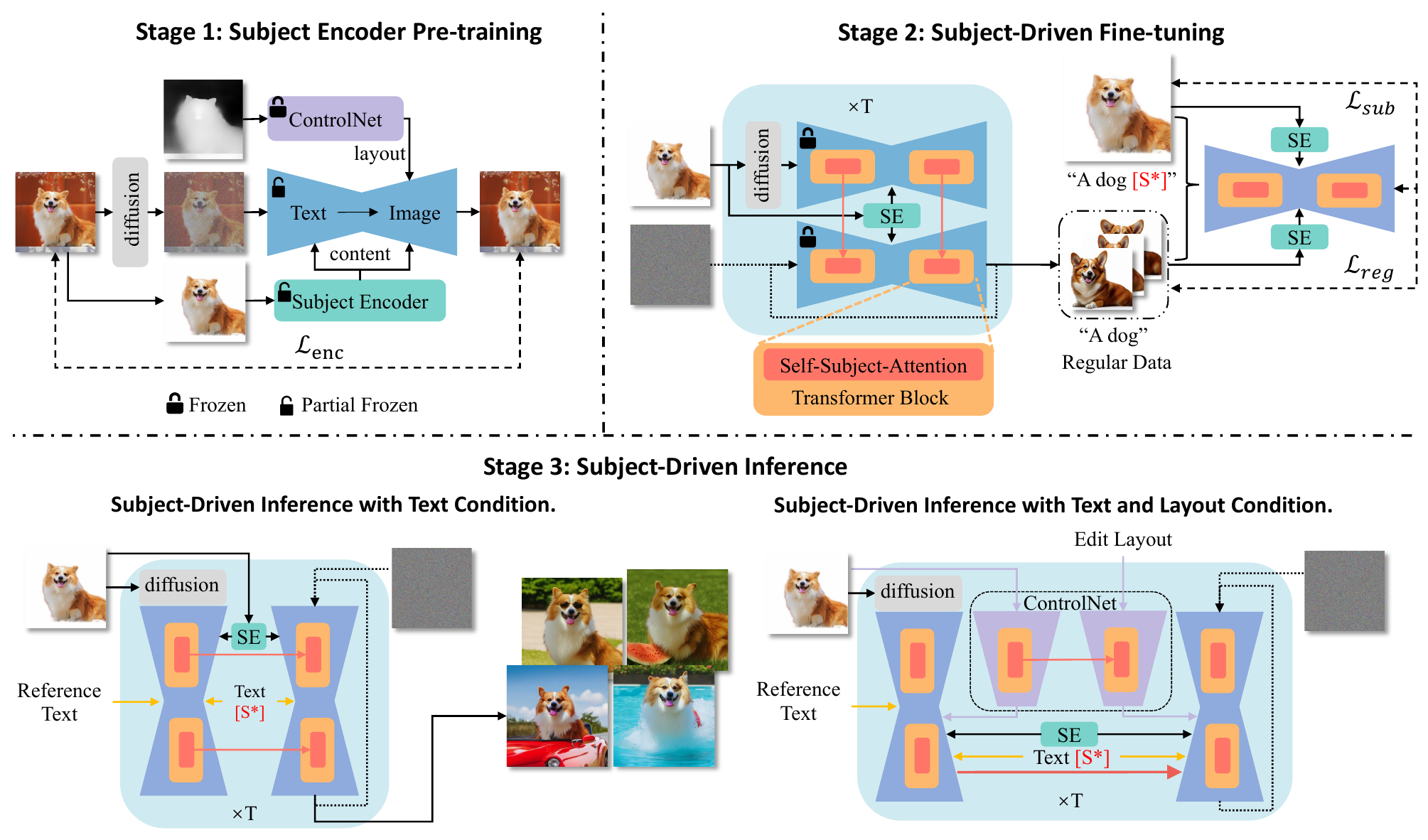}
  \caption{Overview of the proposed DreamTuner framework. Firstly, a subject-encoder (SE) is trained for coarse identity preservation, where a frozen ControlNet is utilized to maintain the layout. 
  Then an additional fine-tuning stage like existing methods is conducted with proposed subject-encoder and self-subject-attention for fine identity preservation.
  Finally a refined subject driven image generation model is obtained which could synthesis high-fidelity images of the specific subject controlled by text or other layout conditions.
  It is worth noting that both of the subject-driven fine-tuning stage and inference stage require only a single reference image.}
  \label{fig:framework}
\end{figure*}

Based on the above motivations, we propose DreamTuner, a novel subject-driven image generation method that injects the reference information from coarse to fine. 
Firstly, a subject-encoder is proposed for coarse identity preservation. The main object of the reference image is compressed by a pretrained CLIP \cite{clip} image encoder. And the compressed features are injected to the T2I model through an additional attention layer before visual-text cross-attention. 
% Besides, we also propose a training method that decouples content and layout for subject-encoder.
% Specifically, a depth-based controlnet \cite{controlnet} is trained along with the subject-encoder to prevent the encoder focus on the layout of the reference image. 
Furthermore, we introduce a training technique that decouples content and layout for the subject-encoder. To achieve this, we add a pretrained depth-based ControlNet \cite{controlnet} to avoid the subject-encoder focusing on the layout of reference image. 
During training, ControlNet is frozen to control the layout of generated images so that the subject-encoder could only need to learn the content of the reference image.
% , and then we use only the stable diffusion part of the ControlNet for subject generation.
% Then we modified the self-attention layers to self-subject-attention for fine identity preservation. 
% Based on the above-mentioned method to inject reference features into self-attention layers, self-subject-attention further introduce the weight and mask strategy for better controlability.
Next, we alter the self-attention layers to self-subject-attention to preserve fine identity. Based on the above-mentioned method to inject reference features into self-attention layers, self-subject-attention further introduces weight and mask strategies for improved control.
% It is worth noting that self-subject-attention is an elegant, effective and training-free method for identity preservation, which can be used plug-and-play during inference. 
% Finally, an additional fine-tuning stage can lead to better identity preservation. 
% With the help of subject-encoder and self-subject-attention, the fine-tuning process requires only a few training steps.
Notably, self-subject-attention is a graceful, efficient, and training-free technique for preserving identity that can be used plug-and-play during inference.	
Lastly, an extra phase of fine-tuning can enhance identity preservation. With the aid of subject-encoder and self-subject-attention, the fine-tuning procedure only needs a few training steps.	

The main contributions are summarized as follows:
\begin{itemize}
  \setlength\itemsep{0.3em}
    \item We propose a novel image encoder and fine-tuning based subject-driven image generation method that could generate high-fidelity images of a specific subject with only a single reference image.
    \item A subject-encoder is proposed for coarse identity preservation, which could also reduce the fine-tuning time cost, along with a content and layout decoupled training method.
    \item A novel self-subject-attention is proposed as an elegant, effective, and plug-and-play method for refined subject identity preservation.
\end{itemize}

\section{Related Works}
\subsection{Text-to-Image Generation}
The development of text-to-image generation has evolved from Generative Adversarial Networks (GANs) \cite{gan} based methods \cite{clip2stylegan,attngan,styleclip,cigan} in early years to Vector Quantization Variational AutoEncoders (VQVAEs) \cite{vqvae,vqvae2} and Transformers \cite{transformer} based methods \cite{dalle,vqgan,cogview,parti}, and more recently to diffusion based methods \cite{ddpm,ddim,dalle2,imagen,ldm,huang2022draw}. 

Diffusion models are probabilistic generative models that learn to restore the data distribution perturbed by the forward diffusion process. 
At the training stage, the data distribution is perturbed by a certain scale of Gaussian noise in different time steps, and the diffusion models are trained to predict the denoised data.
The data distribution with the largest scale of noise tends to standard normal distribution.
Thus at the inference stage, diffusion models could sample high-fidelity data from Gaussian noise step by step.

Large-scale diffusion models \cite{dalle2,imagen,ldm} have achieved state-of-the-art performance on text-to-image generation. 
% 待修改，待扩充
DALLE-2 \cite{dalle2} builds a prior model to transform CLIP \cite{clip} text embeddings to CLIP image embeddings and further generates high-resolution images using large-scale hierarchical U-Nets which consists of a image generation decoder and some image super-resolution models.
IMAGEN \cite{imagen} adopt a larger pretrained text encoder \cite{t5} for better image-text consistency.
LDM \cite{ldm} compresses the images through a pretrained auto-encoder and perform diffusion in the latent space, which leads to efficient high-resolution image generation.
With the help of classifier-free guidance \cite{classifierfree}, high-fidelity images can be generated by these large-scale diffusion based text-to-image models.
However, these models are all designed for general scenarios that are controlled only by text. For subject driven generation that requires subject identity preservation, text is far from enough. 

\subsection{Subject-Driven Generation}
Subject-driven image generation requires the model to understand the visual subject contained in the initial reference images and synthesize totally unseen scene of the demonstrated subjects. Most existing works use a pretrained synthesis network and perform test-time training for each subject. For instance, MyStyle \cite{nitzan2022mystyle} adopts a pretrained StyleGAN for personalized face generation. Later on, some works fine-tune the pretrained T2I models to adapt image generation to a speciﬁc unseen subject. Dreambooth \cite{dreambooth} fine-tunes all the parameters, Texture Inversion \cite{textualinversion} introduces and optimizes a word vector for the new concept, Custom-Diffusion \cite{customdiffusion} only fine-tunes a subset of cross-attention layer parameters. For those fine-tuning based methods, more training steps leads to better identity preservation but undermines the generation capacity of the pretrained model. Another line of works train additional image encoders to inject the reference image information for subject generation. ELITE \cite{elite} and Instantbooth \cite{instantbooth} use two mapping networks to convert the reference images to a textual embedding and additional cross-attention layers for subject generation. Taming Encoder\cite{tamingencoder} adopts an encoder to capture high-level identifiable semantics of subjects. SuTI \cite{suti} achieves personalized image generation by learning from massive amount of paired images generated by subject-driven expert models. However, these methods are trained with weakly-supervised data and performs without any test-time fine-tuning, thus leading to much worse faithfulness than DreamBooth \cite{dreambooth}. Recently, MasaCtrl \cite{MasaCtrl} proposes a mask-guided mutual self-attention strategy similar as our self-subject-attention for text-based non-rigid image synthesis and editing without ﬁne-tuning. Our proposed method combines the strengths of those works, it first coarsely injects the subject information using a subject-encoder, then fine-tunes the model with the novel self-subject-attention for better identity preservation.

\section{Method}
In this paper, we propose DreamTuner as a novel framework for subject driven image generation based on both fine-tuning and image encoder, which maintains the subject identity from coarse to fine. 
As shown in Fig. \ref{fig:framework}, DreamTuner consists of three stages: subject-encoder pre-training, subject-driven fine-tuning and subject-driven inference.
Firstly, a subject-encoder is trained for coarse identity preservation.
Subject-encoder is a kind of image encoder that provides compressed image features to the generation model.
A frozen ControlNet \cite{controlnet} is utilized for decoupling of content and layout.
Then we fine-tune the whole model on the reference image and some generated regular images as in DreamBooth \cite{dreambooth}. 
Note that subject-encoder and self-subject-attention are used for regular images generation to refine the regular data.
At the inference stage, the subject-encoder, self-subject-attention, and subject word [S*] obtained through fine-tuning, are used for subject identity preservation from coarse to fine.
Pretrained ControlNet \cite{controlnet} could also used for layout controlled generation.

\subsection{Subject-Encoder}
\begin{figure}
  \centering
  \includegraphics[width=0.8\linewidth]{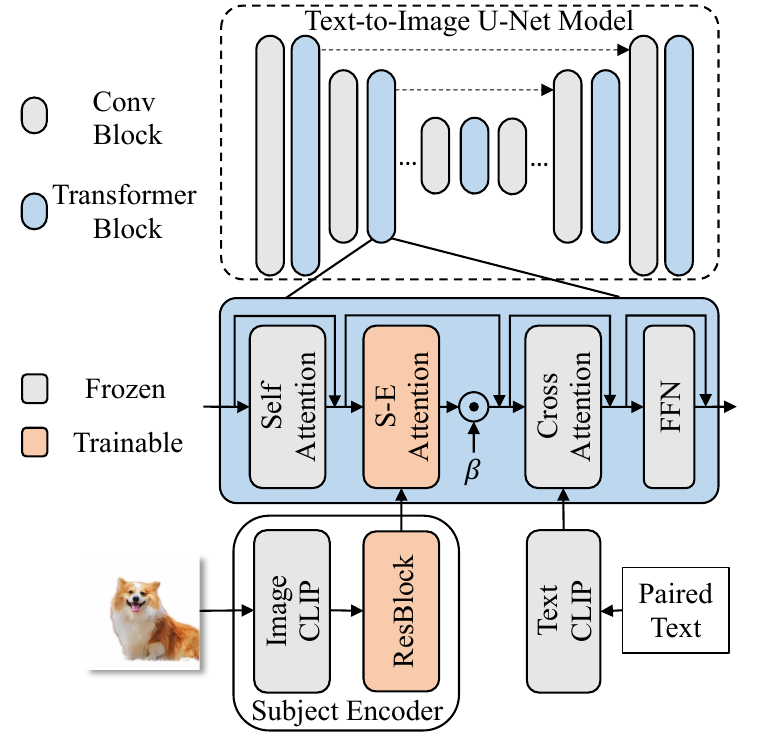}
  \caption{Illustration of the text-to-image generation U-Net model with proposed subject-encoder.}
  \label{fig:subject-encoder}
  % \vspace{-3mm}
\end{figure}
Image encoder based subject driven image generation methods \cite{elite,instantbooth} have proved that the compressed reference image features can guide the pretrained text-to-image model to generate a general appearance of the subject.
Thus we propose subject-encoder as a kind of image encoder that provides a coarse reference for subject driven generation.
As shown in Fig. \ref{fig:subject-encoder}, a frozen CLIP \cite{clip} image encoder is used to extract the compressed features of reference image. 
Salient Object Detection (SOD) model or segmentation model are used to remove the background of the input image and emphasize the subject.
Then some residual blocks (ResBlock) are introduced for domain shift. 
Multi-layer features extracted by CLIP are concatenated in the channel dimension and then adjust to the same dimension as the generated features through the residual blocks. 

The encoded reference features of subject-encoder are injected to the text-to-image model using additional subject-encoder-attention (S-E-A) layers. 
The subject-encoder-attention layers are added before the visual-text cross-attention, because the cross-attention layers are the modules that control the general appearance of generated images.
We build the subject-encoder attention according to the same settings as cross-attention and zero initial the output layers.
An additional coefficient $\beta$ is introduced to adjust the influence of subject-encoder, as in Eq. \ref{eq:beta}.
\begin{equation}
\begin{aligned}
\text{S-E-A}(\mathbf{z},\text{SE}(\mathbf{r}))'=\beta*\text{S-E-A}(\mathbf{z},\text{SE}(\mathbf{r}))
\label{eq:beta}
\end{aligned}
\end{equation}
where $\mathbf{z}$ is the generated image features and $\mathbf{r}$ is the reference image.
It should be noted that only the additional residual blocks and subject-encoder-attention layers are trainable during training to maintain the generation capacity of the text-to-image model.

Besides, we found that the subject-encoder will provide both the content and the layout of the reference image for text-to-image generation.
However, in most cases, layout is not required in subject driven generation.
Thus we further introduce ControlNet \cite{controlnet} to help decouple the content and layout.
Specifically, we train the subject-encoder along with a frozen depth ControlNet, as in Fig. \ref{fig:framework}.
As the ControlNet has provided the layout of reference image, the subject-encoder can focus more on the subject content.

\subsection{Self-Subject-Attention}
\begin{figure}
  \centering
  \includegraphics[width=\linewidth]{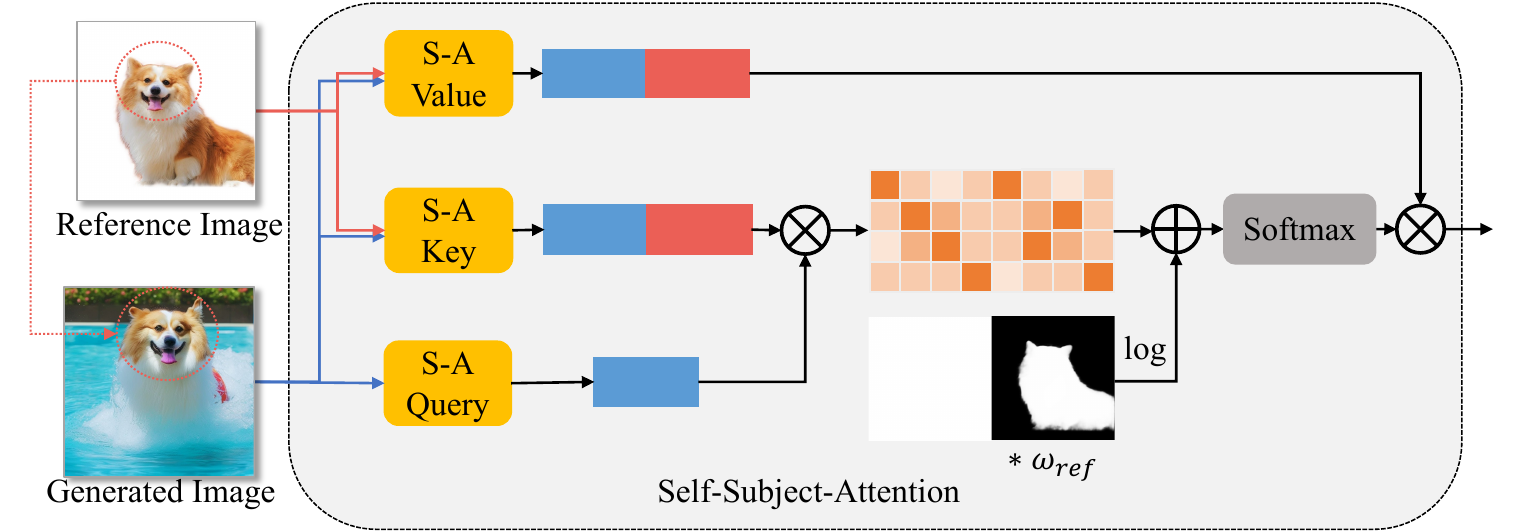}
  \caption{Illustration of the proposed self-subject-attention. S-A indicates self-attention.}
  \label{fig:ss-attn}
  % \vspace{-5mm}
\end{figure}

As discussed in some text-to-video works \cite{tune_a_video,t2v_zero}, self-attention layers in pretrained text-to-image models could be used for content consistency across-frames.
Since the subject-encoder has provided general appearance of the specific subject, we further propose self-subject-attention based on the self-attention layers for fine subject identity preservation.
The features of reference image extracted by the text-to-image U-Net model are injected to the self-attention layers, which can provide refined and detailed reference because they share the same resolution with the generated features.
Specifically, as shown in Fig. \ref{fig:framework} Stage-2 and Stage-3, the reference image are noised through diffusion forward process at time step $t$.
Then the reference features before self-attention layers are extracted from the noised reference image, which share the same data distribution with the generated features at time step $t$.
The original self-attention layers are modified to self-subject-attention layers by utilizing reference features.
As demonstrated in Fig. \ref{fig:ss-attn}, the features of generated image are taken as the query and the concatenation of generated features and reference features is taken as the key and value.
To eliminate the influence of reference image background, SOD model or segmentation model are used to create a foreground mask, which uses 0 and 1 to indicate the background and foreground.
Besides, the mask can also be used to adjust the scale of the impact of reference image through a weight strategy, i.e., multiply the mask by an adjustment coefficient $\omega_{ref}$.
The mask works as an attention bias, thus a log function is used as a preprocessing. 
It is worth noting that the extraction of reference features is unrelated to the generated image, so additional reference text can be used to enhance consistency with the reference image.

The original Self-Attention (S-A) is formulated as Eq. \ref{eq:sa}.
\begin{equation}
\begin{aligned}
\text{S-A}(\mathbf{z})&=Softmax(\frac{\mathbf{q}\cdot \mathbf{k}^T}{\sqrt{d}})\cdot \mathbf{v}\\
w.r.t.\quad &\mathbf{q}=Query(\mathbf{z}),\mathbf{k}=Key(\mathbf{z}),\mathbf{v}=Value(\mathbf{z})
\label{eq:sa}
\end{aligned}
\end{equation}
where $Query,Key,Value$ are linear layers in self-attention, $d$ is the dimension of $\mathbf{q},\mathbf{k},\mathbf{v}$. 
The proposed Self-Subject-Attention (S-S-A) is formulated as Eq. \ref{eq:ssa}.
\begin{equation}
\begin{aligned}
\text{S-S-A}(\mathbf{z},\mathbf{z}^r)&=Softmax(\frac{\mathbf{q}\cdot\left[\mathbf{k},\mathbf{k}^r\right]^T}{\sqrt{d}}+\log \mathbf{m})\cdot\left[\mathbf{v},\mathbf{v}^r\right]\\
w.r.t.\quad &\mathbf{k}^r=Key(\mathbf{z}^r),\mathbf{v}^r=Value(\mathbf{z}^r)\\
& \mathbf{m}=\left[\mathbf{J},\omega_{ref}*\mathbf{m}^r\right]
\end{aligned}
\label{eq:ssa}
\end{equation}
where $\mathbf{z}^r$ is the reference image features, $\left[\right]$ indicates concatenation, $\mathbf{m}^r$ is the foreground mask of reference image, $\mathbf{J}$ is a all-ones matrix.
As in Re-Imagen \cite{reimagen}, there are more than one conditions in DreamTuner, thus we modify the original classifier free guidence to a interleaved version: 
\begin{equation}
\begin{aligned}
\mathbf{\hat{{\epsilon}}}_t^r&=\omega_{r}*{\mathbf{\epsilon}}_{\theta}(\mathbf{x}_t,\mathbf{c},\mathbf{x}^r_{t-\Delta t},t)-(\omega_{r}-1)*{\mathbf{\epsilon}}_\theta(\mathbf{x}_t,\mathbf{c},\emptyset,t)
\label{eq:cf1}
\end{aligned}
\end{equation}
\begin{equation}
\begin{aligned}
\mathbf{\hat{{\epsilon}}}_t^c&=\omega_c*\mathbf{\epsilon}_\theta(\mathbf{x}_t,\mathbf{c},\mathbf{x}^r_{t-\Delta t},t)-(\omega_c-1)*\mathbf{\epsilon}_\theta(\mathbf{x}_t,\mathbf{uc},\mathbf{x}^r_{t+\Delta t'},t)
\label{eq:cf2}
\end{aligned}
\end{equation}
\begin{equation}
\begin{aligned}
\hat{\mathbf{\epsilon}}_t&=\left\{\begin{aligned}
\hat{\mathbf{\epsilon}}_t^r,&\quad \lambda_t\leq p_{r}\\
\hat{\mathbf{\epsilon}}_t^c,&\quad \lambda_t>p_{r}
\end{aligned}\right.\lambda_t\sim{U}(0,1)
\end{aligned}
\end{equation}
where $\mathbf{x}_t$ is the generated image at time step $t$, $\mathbf{c}$ is the condition, $\mathbf{uc}$ is the undesired condition, $\omega_{r}$ and $\omega_{c}$ are the guidance scales of reference image and text condition.
$\mathbf{x}^r_{t-\Delta t}$ and $\mathbf{x}^r_{t+\Delta t'}$ are the diffusion noised reference images at time step $t-\Delta t$ and $t+\Delta t'$, $\Delta t$ and $\Delta t'$ are small time step biases which are used for the adjustment of noise intensity of reference image. 
$\hat{\mathbf{\epsilon}}_\theta$ is the diffusion model and $\theta$ represents the parameters. $\hat{\mathbf{\epsilon}}_t$ is the final output at time step $t$.
Eq. \ref{eq:cf1} emphasizes the guidance of reference image and Eq. \ref{eq:cf2} emphasizes the guidance of condition, where $p_{r}$ controls the possibility of selecting Eq. \ref{eq:cf1}.

\subsection{Subject Driven Fine-tuning}
While excellent performance can be achieved through SE and S-S-A, a few additional fine-tuning steps leads to better subject identity preservation. 
As shown in Fig. \ref{fig:framework} Stage-2, We conduct the fine-tuning stage similar to DreamBooth \cite{dreambooth}, while other fine-tuning methods, such as LoRa \cite{hu2021lora}, also works.
Firstly, some regular images are generated to help the model learn the specific identity of the subject and maintain the generation capacity.
The paired data are built as \{subject reference image, "A \emph{class\_word} [S*]"\} and \{regular image, "A \emph{class\_word}"\}. 
Then the paired data is used to fine-tune the pretrained text-to-image model. 
All of the parameters including CLIP text encoder are trainable in this stage as in DreamBooth. 
In order to achieve more refined subject learning, we make four improvements:
\begin{enumerate}
    \item The background of reference image is replaced by a simple white background to focus on the subject itself.
    \item The word [S*] is an additional trainable embedding as in Textual Inversion \cite{textualinversion} rather than a rare word \cite{dreambooth}.
    \item The subject-encoder is trained with the text-to-image generation model for better subject identity preservation.
    \item To learn the detailed identity of the subject, we generate regular images that are more similar to the reference image by using subject-encoder, self-subject-attention and detailed caption of the reference image. The model are required to learn the subject details to distinguish "A \emph{class\_word}" and "A \emph{class\_word} [S*]"
\end{enumerate}
\begin{figure*}
  \centering
  \includegraphics[width=\linewidth]{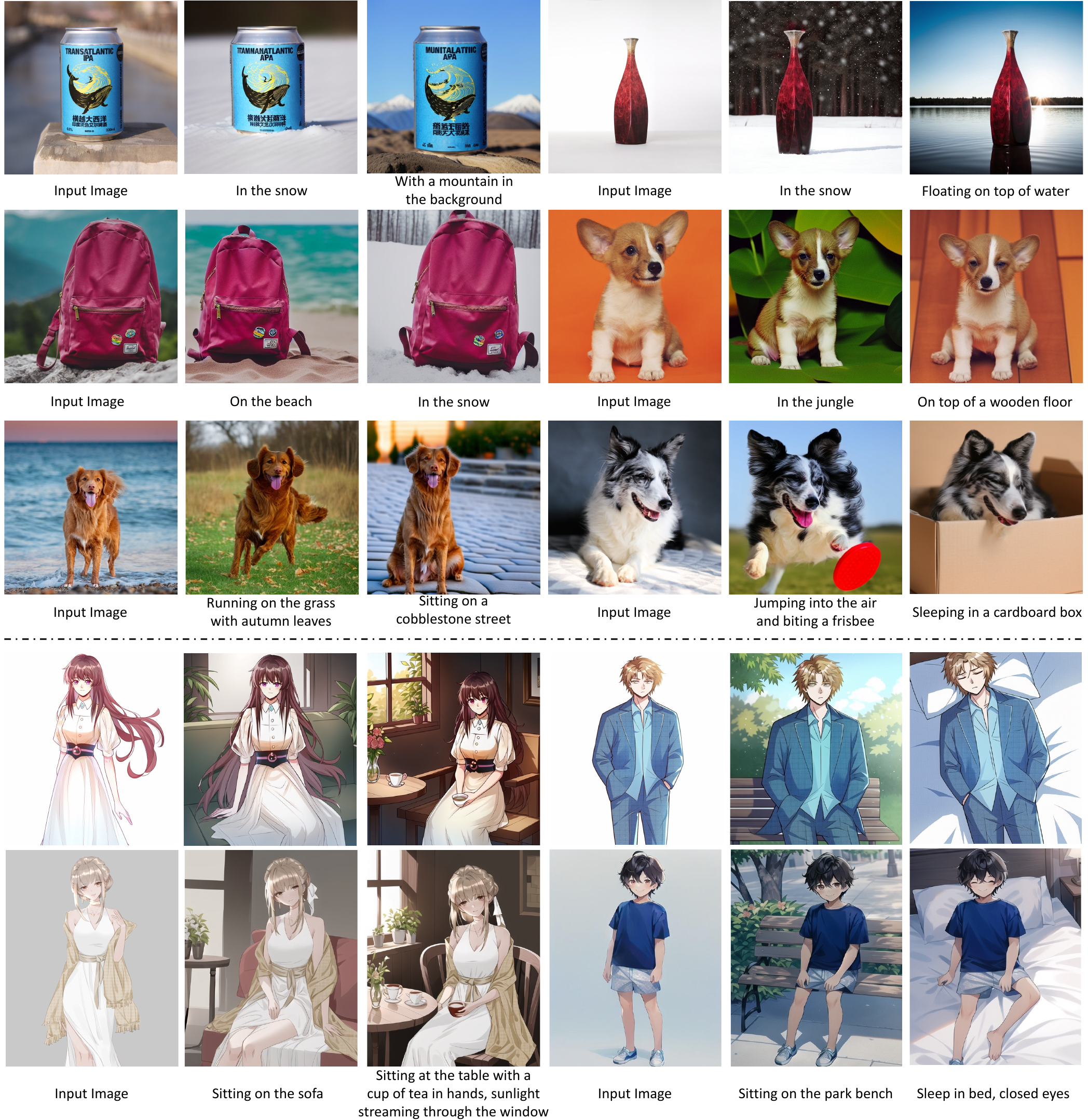}
  \caption{Subject driven generation results of DreamTuner.}
  \label{fig:results}
  \vspace{-1mm}
\end{figure*}
\begin{figure*}
  \centering
  \includegraphics[width=\linewidth]{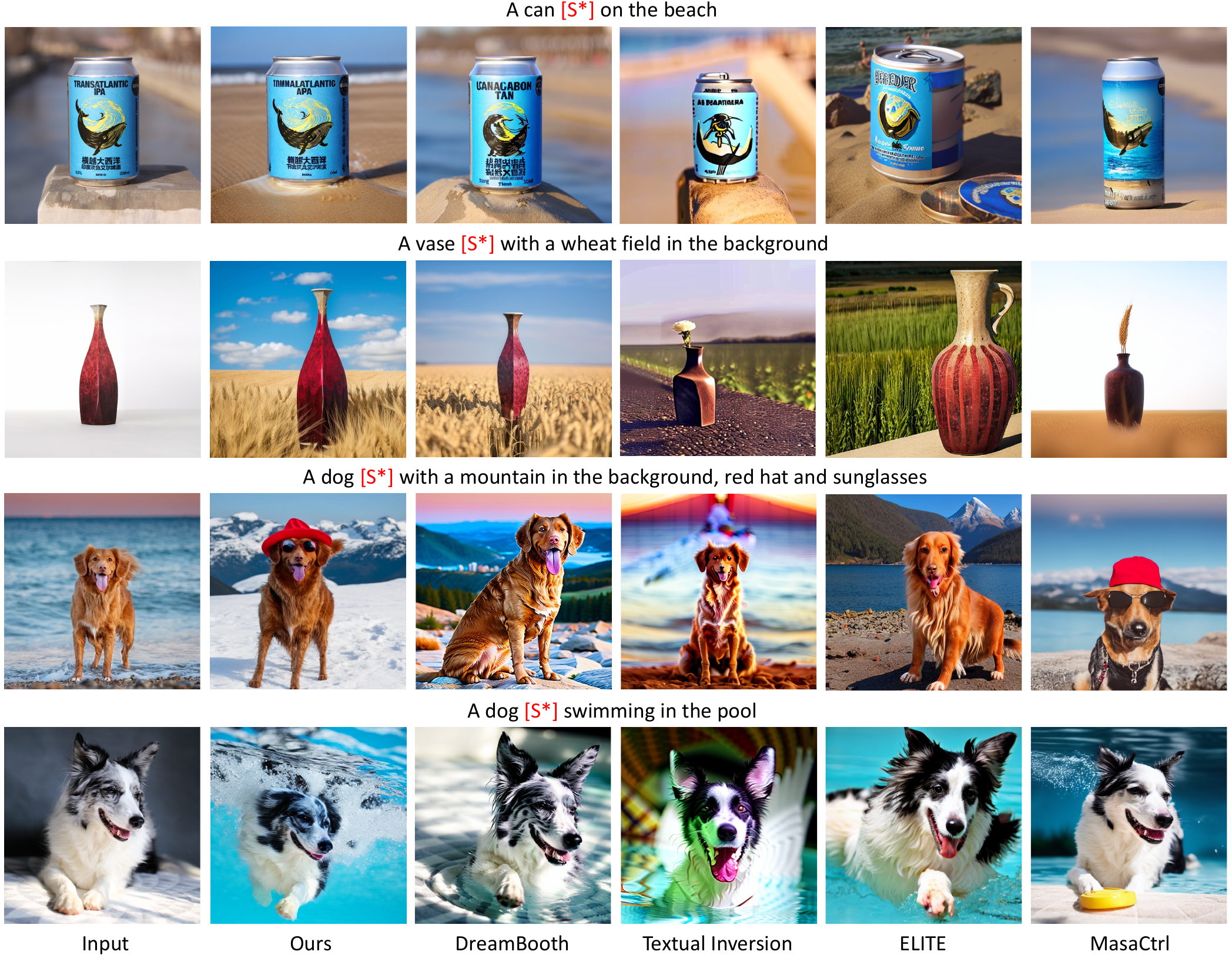}
  \caption{Comparison between our proposed DreamTuner and state-of-the-art subject driven image generation methods.}
  \label{fig:comparison}
  \vspace{-1mm}
\end{figure*}
\section{Experiments}

\subsection{Implementation Details}
We build our DreamTuner based on the Stable Diffusion model \cite{ldm}.
For the subject-encoder, CLIP features from layers with layer indexes \{25, 4, 8, 12, 16\} are selected as in ELITE \cite{elite}.
Elastic transform and some other data augmentation methods are used to enhance the difference between input reference images and the generated images.
InSPyReNet \cite{InSPyReNet} is adopted as the SOD model for natural images and a pretrained anime segmentation model is used to separate the foreground and background for anime character images.
For self-subject-attention, $p_r$ is set to 0.9, $\omega_{ref}$ is set to 3.0 for natural images and 2.5 for anime images. 
% $\Delta t$ and $\Delta t'$ are implemented by scaling the added noise by 0.99 and 1.01. 
The training steps of subject-driven fine-tuning is set to 800-1200.
BLIP-2 \cite{blip2} and deepdanbooru \cite{deepdanbooru} are used as the caption models for natural images and anime character images.
More details could be found in Appendix-\ref{appendix:results}.

\subsection{Subject Driven Generation Results}
Subject driven generation results could be found in Fig. \ref{fig:results}, consisting of static objects, animals, and anime characters.
It could be found that the subject identities have been preserved well and the generated images are consistent with the text condition.
The subject appearance is maintained even with some complex texts.
It is worth noting that DreamTuner could achieve excellent identity preservation on details, e.g., the words on the can in the first row, the white stripes of the dog in the second raw, the eyes and clothes of the anime characters, etc. 
% More results along with the pose controlled generation results could be found in the project page.
% \footnote{\url{https://dreamtuner-diffusion.github.io/}}.
\begin{table}
  \caption{Subject fidelity (CLIP-I) and prompt fidelity (CLIP-T) quantitative metric comparison. The top-2 results of each metric have been emphasized.}
  \label{tab:comparison}
  \centering
  \begin{tabular}{c|cc}
    \toprule
    Method & CLIP-T & CLIP-I\\
    \midrule
    Textual Inversion & 0.251 & 0.722  \\
    Masactrl & \textbf{0.296} & 0.735  \\
    ELITE & 0.271 & 0.726  \\
    DreamBooth & 0.267 & \textbf{0.788}  \\
    \midrule
    DreamTuner & \textbf{0.281} & \textbf{0.767}  \\
  \bottomrule
\end{tabular}
\vspace{-3mm}
\end{table}
\begin{figure*}[htbp]
\subfloat{
\begin{minipage}[t]{0.16\linewidth}
% \vspace{0.0025\linewidth}
  % \scriptsize{"a photo of girl [S*], \\sitting at}\\ 
  % \scriptsize{a desk, typing on a computer"}\\ 
  \centering\includegraphics[width=.99\linewidth]{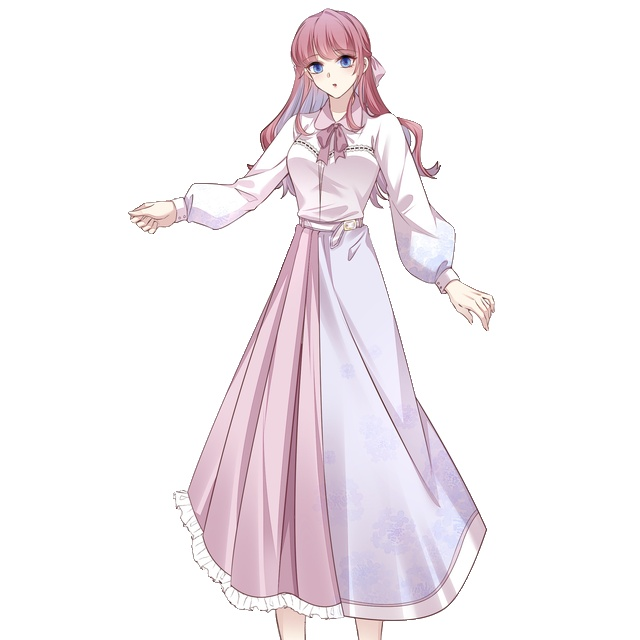}
  \centerline{\scriptsize{Input}}\\
  \centering\includegraphics[width=.99\linewidth]{pics/ablation/input.png}
  \centerline{\scriptsize{Input}}
\end{minipage}
}
\subfloat{
\begin{minipage}[t]{0.16\linewidth}
    \includegraphics[width=.99\linewidth]{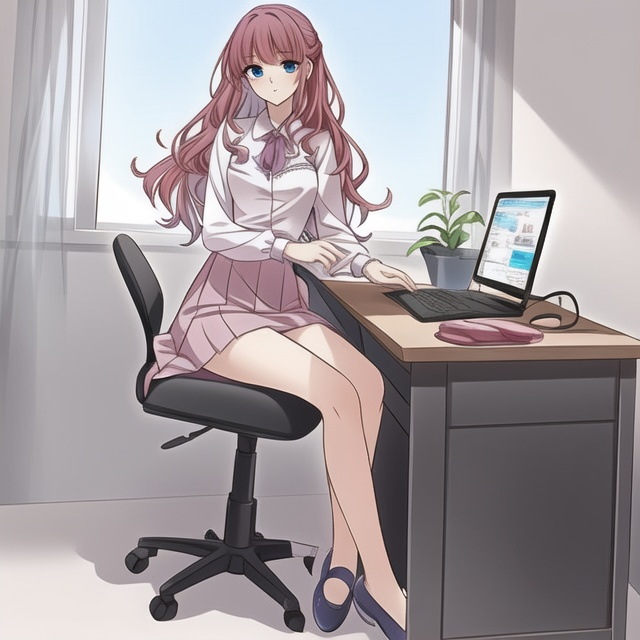}
    \centering\scriptsize{$\beta$=0.0}\\
    \includegraphics[width=.99\linewidth]{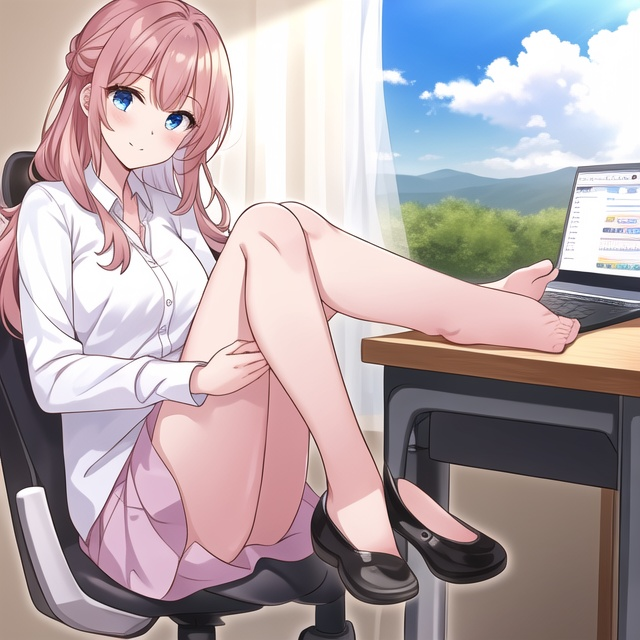}
    \centering\scriptsize{$\omega_{ref}$=0.0}
\end{minipage}
\begin{minipage}[t]{0.16\linewidth}
    \includegraphics[width=.99\linewidth]{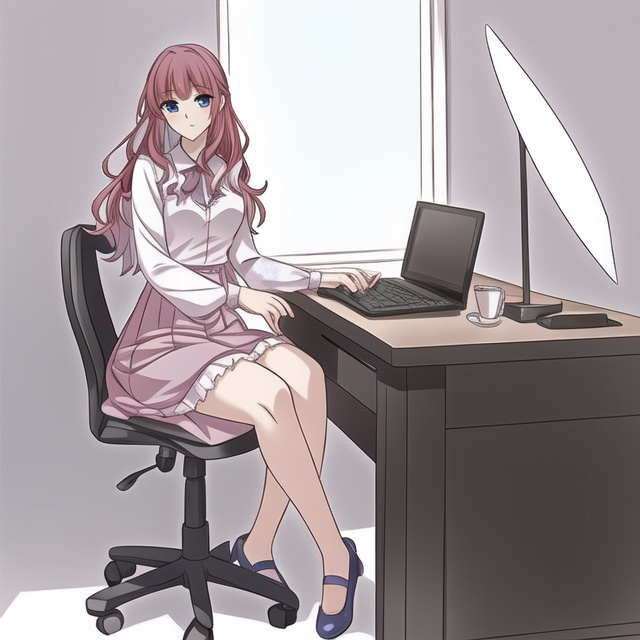}
    \centering\scriptsize{$\beta$=0.1}\\
    \includegraphics[width=.99\linewidth]{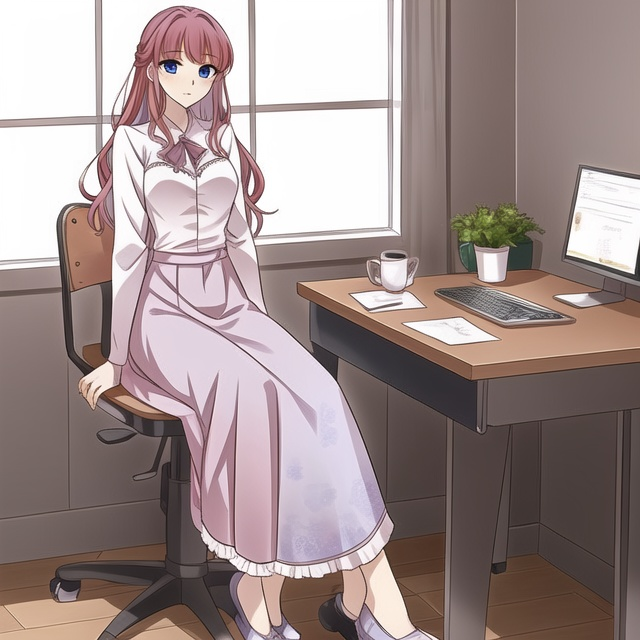}
    \scriptsize{$\omega_{ref}$=1.0}
\end{minipage}
\begin{minipage}[t]{0.16\linewidth}
    \includegraphics[width=.99\linewidth]{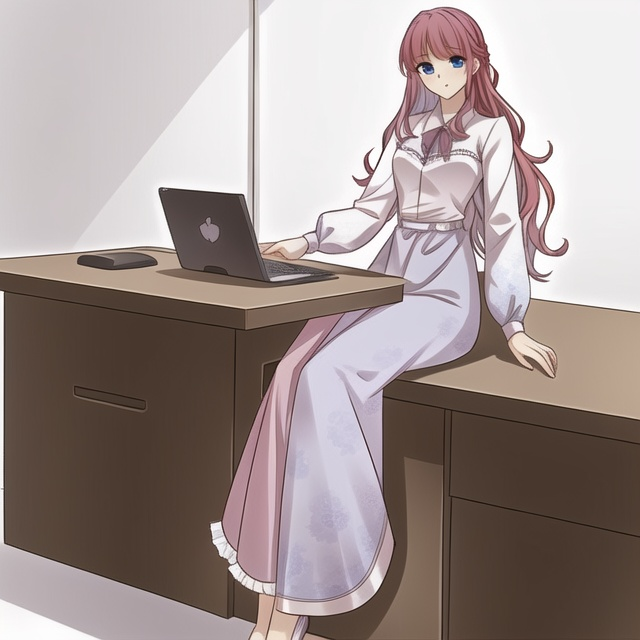}
    \centering\scriptsize{$\beta$=0.2} \\
    \includegraphics[width=.99\linewidth]{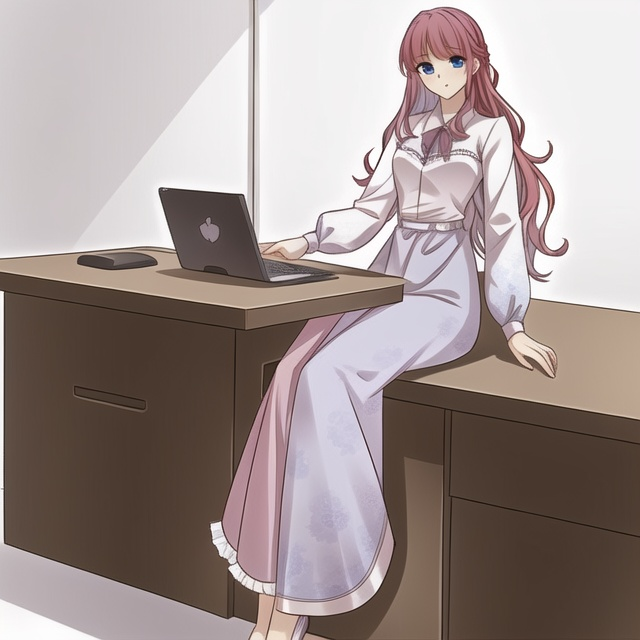}
    \centering\scriptsize{$\omega_{ref}$=2.5}
\end{minipage}
\begin{minipage}[t]{0.16\linewidth}
    \includegraphics[width=.99\linewidth]{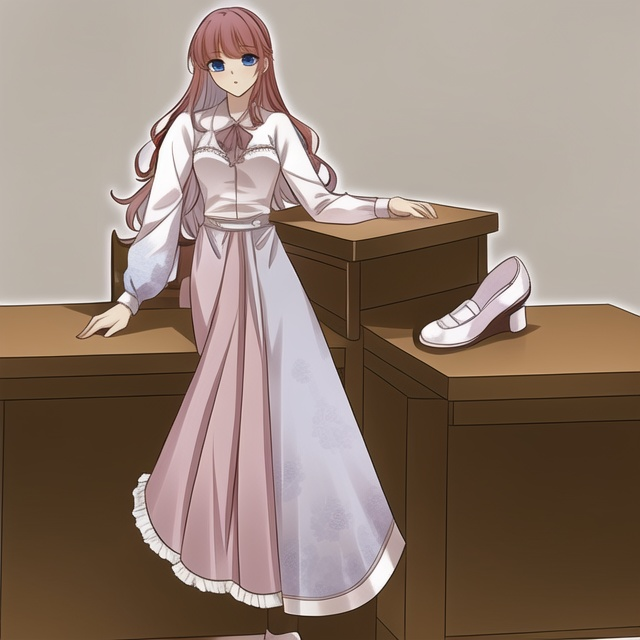}
    \centering\scriptsize{$\beta$=0.5}\\
    \includegraphics[width=.99\linewidth]{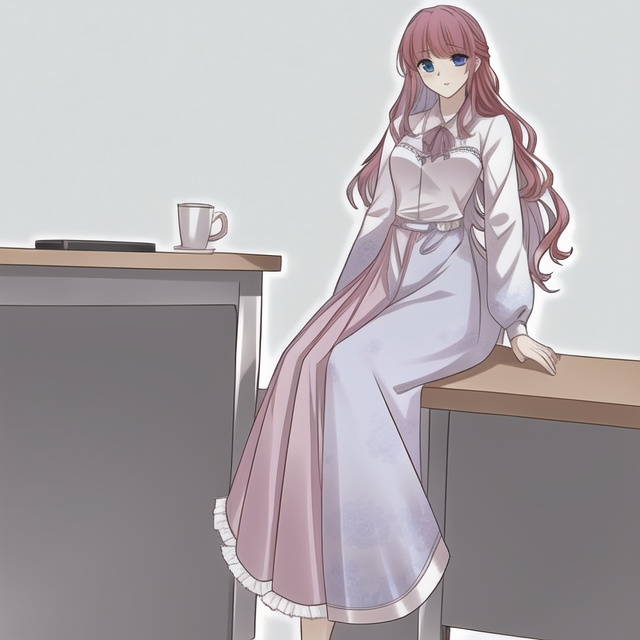}
    \centering\scriptsize{$\omega_{ref}$=10.0}
\end{minipage}
\begin{minipage}[t]{0.16\linewidth}
    \includegraphics[width=.99\linewidth]{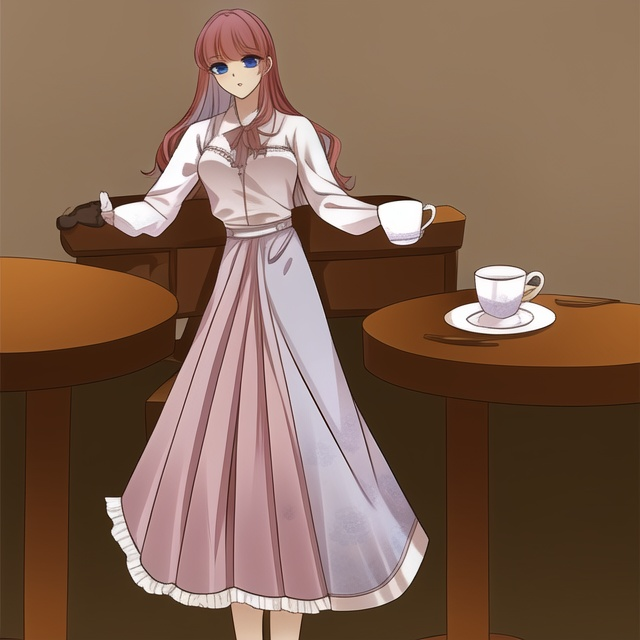}
    \centering\scriptsize{SE without ControlNet}\\
    \includegraphics[width=.99\linewidth]{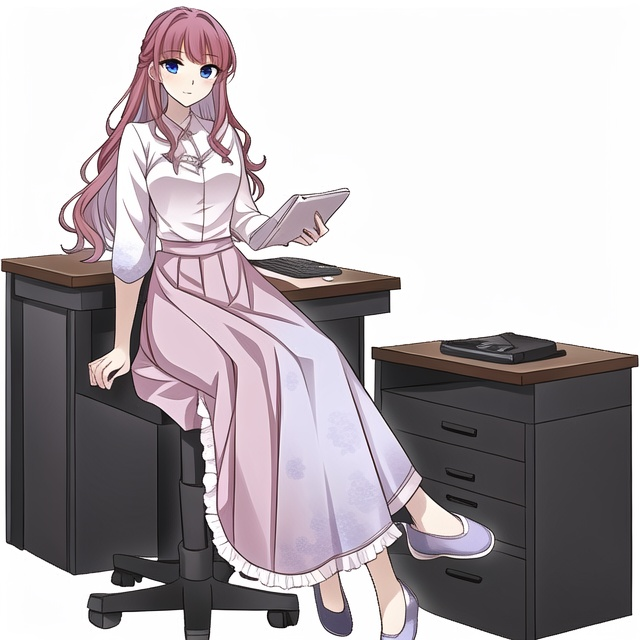}
    \centering\scriptsize{S-S-A without mask}
\end{minipage}
}
\caption{Ablation study of subject-encoder and self-subject-attention. The input text is 'a photo of girl [S*], sitting at a desk, typing on a computer'}
\label{fig:ablation}
\vspace{-1mm}
\end{figure*}
\subsection{Comparison with Other Methods}
The comparison results between our proposed DreamTuner and other subject driven image generation methods are shwon in Fig. \ref{fig:comparison}.
The corresponding quantitative metric comparison could be found in Table. \ref{tab:comparison}.
We use CLIP-T, the average cosine similarity between CLIP image embeddings and CLIP text embeddings, as the metric to evaluate the consistency between generated images and input text prompts.
CLIP-I, the average cosine similarity between CLIP embeddings of generated images and reference images, is adopted as a metric that indicates the subject fidelity. 
Only a single image are used as the reference image for each method.  
It could be found that our proposed DreamTuner outperforms the fine-tuning based methods (DreamBooth \cite{dreambooth} and Textual Inversion \cite{textualinversion}), image encoder based method (ELITE \cite{elite}) and improved self-attention based method (MasaCtrl \cite{MasaCtrl}). 
DreamBooth \cite{dreambooth} performs best in CLIP-I but not well on CLIP-T, as it is easy to overfit on such a single reference image with only 1200 steps, and less steps leads to worse subject fidelity.
Besides, it could be found in Fig. \ref{fig:comparison} that DreamTuner can generate better subject details benefiting from subject-encoder and self-subject-attention.
Similar to DreamBooth, Textual Inversion \cite{textualinversion} trained with only a single image also has difficulties in the trade-off between avoid overfitting and identity preservation.
Compared with ELITE \cite{elite}, DreamTuner use additional self-subject-attention and subject-driven fine-tuning that leads to better identity preservation.
MasaCtrl \cite{MasaCtrl} shows the best CLIP-T because it does not train the text-to-image model and inherits the creative power of the pretrained model. 
However, it only use a improved self-attention for identity preservation while our proposed DreamTuner introduces subject-encoder and subject-driven fine-tuning, thus achieves better subject fidelity.

\subsection{Ablation Study}
We conduct ablation studies to evaluate the effects of various components in our method, including the subject-encoder and the self-subject-attention. Fig. \ref{fig:ablation} shows the input reference image, corresponding prompt, and results of different settings.

\noindent\paragraph{\textbf{Effect of Subject-Encoder.}} For subject-encoder, $\beta$ is introduced to control how much subject information is injected into the model. To evaluate its effect, we vary its value from $0.0$ to $0.5$, and the generated results are shown in the first row of Fig. \ref{fig:ablation}. From the figure, we see that with the increasing of $\beta$, the consistency between the synthesized image and generated image is improved. However, when the value of $\beta$ is too large(i.e., $0.5$), it may lead to degenerated editing results. As a result, for a tradeoff between inversion and editability, we set $\beta=0.2$. We found that this setting works well for most cases. We also conduct studies of training subject-encoder without ControlNet \cite{controlnet}. In this case, the subject-encoder cannot decouple the content and layout properly, therefore the generated images may have similar layout as the reference images, rather than matching the input prompt.

\noindent\paragraph{\textbf{Effect of Self-Subject-Attention.}} We further analyze the effect of the proposed self-subject-attention. From the images in the last row of Fig. \ref{fig:ablation}, we observe a similar phenomenon when increasing the $\omega_{ref}$ from $0.0$ to $10.0$. When $\omega_{ref}$ is too small, such as close to $0.0$, the generated image is not similar to the input image, and a large $\omega_{ref}$ value may degrade the quality of the generated image. 
When using S-S-A without mask, i.e., $\mathbf{m}^r$ is set to an all-ones matrix, the white background of reference image will also play a significant role and do harm to the generated image.
More ablations are provided in Appendix-\ref{appendix:ablation}.
\section{Conclusion}

In this paper, we focus on the task of subject driven image generation with only a single reference image. Our method, DreamTuner, combines the strengths of existing fine-tuning and additional image encoder-based methods. Firstly, we use a subject-encoder for coarse subject identity preservation. Secondly, we fine-tune the model with self-subject-attention layers to refine the details of the target subject. This allows us to generate high-fidelity images based on only one reference image, with fewer training steps. Furthermore, our method is highly flexible in editing learned concepts into new scenes, guided by complex texts or additional conditions such as pose. This makes it an invaluable tool for personalized text-to-image generation.

{
    \small
    \bibliographystyle{ieeenat_fullname}
    \bibliography{main}
}
\appendix
\section{More Implementation Details}
In this section, we will provide more implementation details.

For the pretraining of subject-encoder, we use the pretrained OpenCLIP \footnote{\url{https://github.com/mlfoundations/open_clip}} model to extract multilayer image features. We selected the features from indices \{$25$, $4$, $8$, $12$, $16$\}. We then train our encoder on two datasets: the natural images dataset, which is a sub-set of LAION-5B \cite{laion-5b}, and the anime character images dataset, which is selected from Danbooru \cite{Danbooru2021}. For natural images, we use the V1-5 version of Stable Diffusion, and for anime images, we use the Anything V3 model \footnote{\url{https://huggingface.co/Linaqruf/anything-v3.0}}. We add a pretrained depth-based ControlNet \cite{controlnet} to make the subject-encoder focus on content since layout is controlled by the ControlNet. To enhance the robustness of the subject-encoder, we use data augmentation techniques such as elastic transform, random flip, blur, and random scale and rotation. 
We set $\beta=1.0$ during the training process and use a batch size of $16$ with a learning rate of $1e-5$. We conduct all experiments on $4$ A100 GPUs, and the training time for each experiment is approximately one week.

At the subject-driven fine-tuning stage, we generate 32 regular images for each subject. Subject-encoder and self-subject-attention are used to generate regular images that are similar to the reference image, where $\beta$ is set to 0.2, $\omega_{ref}$ is set to 3.0 for natural images and 2.5 for anime images. 
Mask strategy is not used in self-subject-attention when generating regular images, as the background is also expected to be maintained.
The U-Net and CLIP \cite{clip} text encoder in the Stable Diffusion model , and the ResBlocks in the subject-encoder are trainable during fine-tuning. The CLIP image encoder is still frozen.
To use a specific word to represent the subject, we train an additional word embedding [S*], which is initialized by the class word. 
The learning rate of the [S*] word embedding is set to 5e-3 and the learning rate of other parameters is set to 1e-6. 
The batch size is set to 2, consisting of a reference image and a regular image. Flip with a probability of 0.5 is adopted as a data augmentation method. We fine-tune the model for 800-1200 steps. It is worth noting that 800 steps are enough for most subjects.
\begin{figure*}
  \centering
  \includegraphics[width=0.9\linewidth]{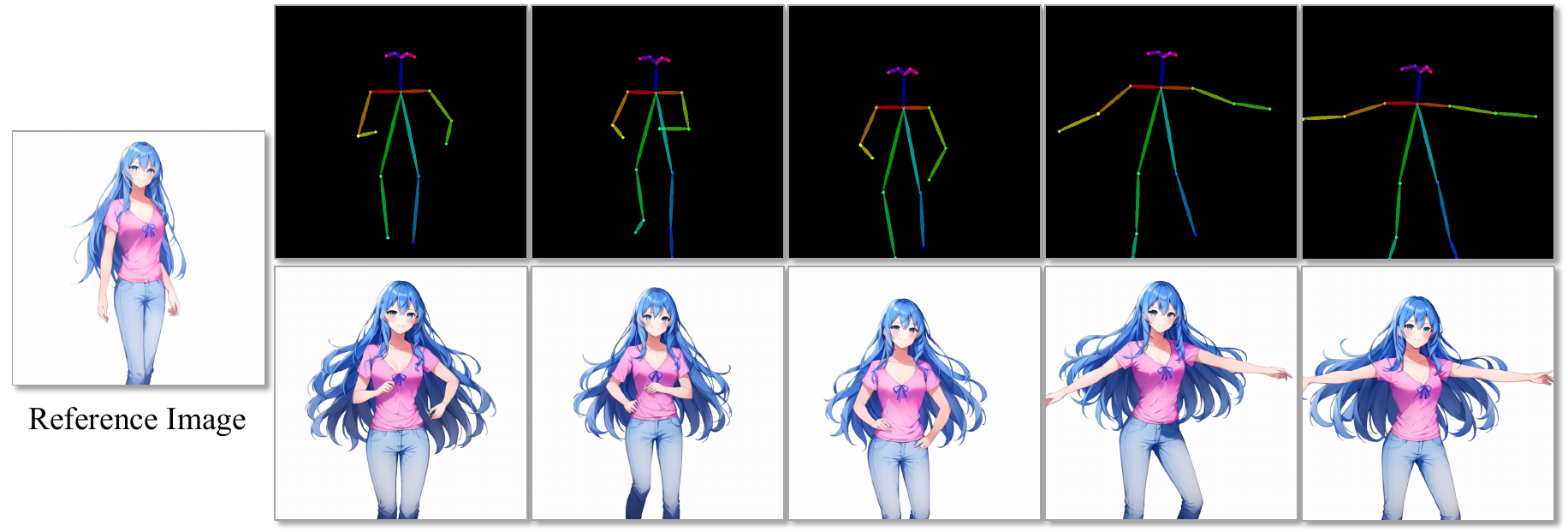}
  \caption{Text and pose controlled subject-driven video generation results.}
  \label{fig:video}
\end{figure*}
\section{More Generated Results}
\label{appendix:results}
In the paper, we have shown the generated results guided by text descriptions.
And the pretrained ControlNets \cite{controlnet} could also be used for DreamTuner through a simple model transfer process \cite{controlnet}. 
As shown in Fig. \ref{fig:pose}, with an additional pose version ControlNet, DreamTuner can generate high fidelity images of the reference anime character, which share the same pose as the input conditions.
Besides, the text prompts can still play its role.
For example, the background could be controlled by the text descriptions, such as simple white background, road, etc.
The expression of the generated characters could also be controlled by the text, such as closed eyes.
Additional objects like flowers and white wings could also be added through text prompts. 
Besides, DreamTuner could also be used for pose controlled subject-driven video generation, as shwon in Fig. \ref{fig:video}.
We also use the first frame and the previous frame as the reference images in self-subject-attention for cross-frame coherence.
The generated video could be found in the project page\footnote{\url{https://dreamtuner-diffusion.github.io/}}.

\section{Ablation Study of Fine-tuning}
\label{appendix:ablation}
We conduct ablation studies on fine-tuning stage to evaluate the effects of subject-encoder and self-subject-attention, as depicted in Fig. \ref{fig:ablation}. The figure displays the input reference image, corresponding prompt, and results of various settings. The first row presents the results of original DreamBooth \cite{dreambooth} achieved at different training steps. The second row exhibits results of DreamTuner without subject-encoder at the fine-tuning stage, while the last row shows the results of integral DreamTuner. The figure elucidates that our method generates images that resemble the input reference image and comply well with the target prompt. Additionally, our method converges faster than the original DreamBooth, which benefits from the subject-encoder. At approximately $300$ fine-tuning steps our method produces satisfactory results.
\begin{figure*}
  \centering
  \includegraphics[width=\linewidth]{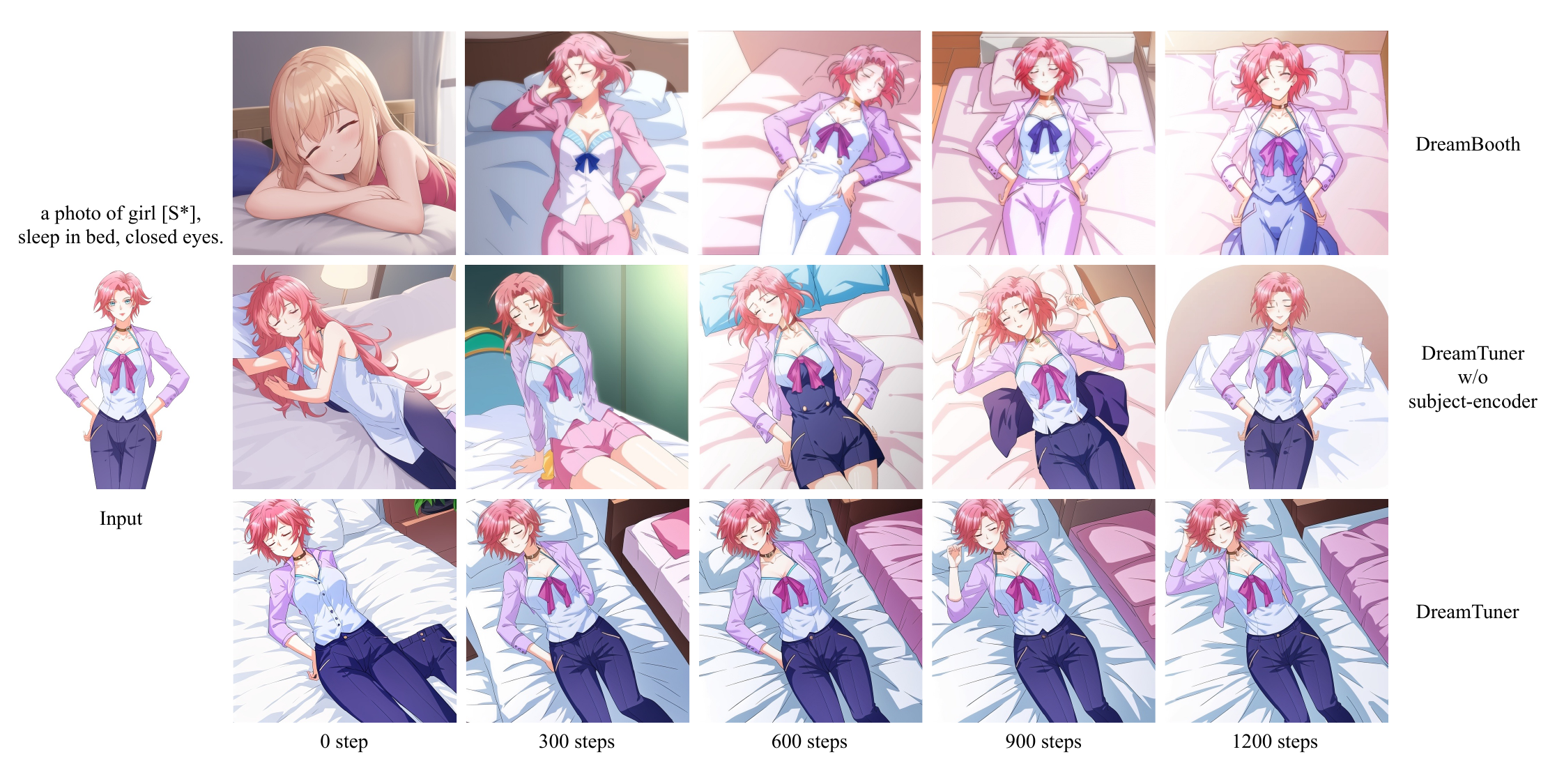}
  \caption{Ablation study at the subject-driven fine-tuning stage.}
  \label{fig:ablation}
\end{figure*}

\end{document}